\documentclass[10pt,twocolumn,letterpaper]{article}

\usepackage[pagenumbers]{cvpr} %
\usepackage[accsupp]{axessibility} %
\usepackage[dvipsnames]{xcolor}

\newcounter{alphasect}
\def\alphainsection{0}

\let\oldsection=\section
\def\section{%
  \ifnum\alphainsection=1%
    \addtocounter{alphasect}{1}
  \fi%
\oldsection}%

\renewcommand\thesection{%
 \ifnum\alphainsection=1%
   \Alph{alphasect}%
 \else
   \arabic{section}%
 \fi%
}%

\usepackage[inkscapelatex=false]{svg}

\definecolor{cvprblue}{rgb}{0.21,0.49,0.74}
\usepackage[pagebackref,breaklinks,colorlinks,citecolor=cvprblue]{hyperref}
\usepackage{xcolor}

\newcommand\blfootnote[1]{%
  \begingroup
  \renewcommand\thefootnote{}\footnote{#1}%
  \addtocounter{footnote}{-1}%
  \endgroup
}

\usepackage{multicol}
\usepackage{multirow}

\def\paperID{10422} %
\def\confName{CVPR}
\def\confYear{2024}

\title{Geometry Transfer for Stylizing Radiance Fields}
\vspace{-11mm}

\author{Hyunyoung Jung$^{1*}$ \qquad
Seonghyeon Nam$^2$
\qquad
Nikolaos Sarafianos$^2$
\qquad
\\
Sungjoo Yoo$^1$
\qquad
Alexander Sorkine-Hornung$^2$
\qquad
Rakesh Ranjan$^2$
\vspace{-3mm}
\\\\
$^1$Seoul National University \qquad $^2$Meta Reality Labs
\\
{\small\bf \url{https://hyblue.github.io/geo-srf}}
}

\begin{document}

\twocolumn[{%
\renewcommand\twocolumn[1][]{#1}%
\maketitle
    \vspace{-20pt}
    \centering
    \captionsetup{type=figure}
    \includesvg[width=\linewidth]{figures/Teaser.svg}

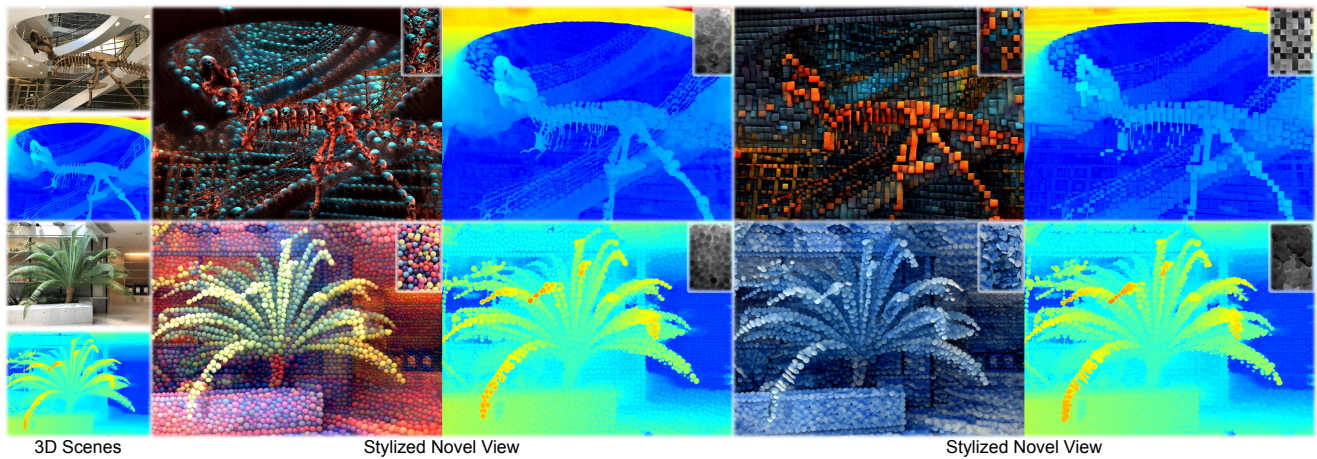
    \captionof{figure}{Given a reference 3D scene and a pair of style guides: an RGB image and a depth map, we coherently stylize both the scene's appearance and shape to best express the given style.}
    \label{fig:teaser}
    \vspace{8pt}
}]
\begin{abstract}
\vspace{-2mm}
Shape and geometric patterns are essential in defining stylistic identity. However, current 3D style transfer methods predominantly focus on transferring colors and textures, often overlooking geometric aspects. In this paper, we introduce Geometry Transfer, a novel method that leverages geometric deformation for 3D style transfer. This technique employs depth maps to extract a style guide, subsequently applied to stylize the geometry of radiance fields. Moreover, we propose new techniques that utilize geometric cues from the 3D scene, thereby enhancing aesthetic expressiveness and more accurately reflecting intended styles. Our extensive experiments show that Geometry Transfer enables a broader and more expressive range of stylizations, thereby significantly expanding the scope of 3D style transfer. 

\blfootnote{*This work was conducted during an internship at Meta}
\end{abstract}
  
\vspace{-0.5cm}
\section{Introduction}\label{sec:intro}
\vspace{-1mm}

With the increasing demand for content creation for virtual and augmented reality, style transfer~\cite{Gatys2016image} has emerged as an innovative technique that bridges the beauty of art with the precision of technology. At its core, style transfer involves rendering one image in the stylistic manner of another, producing a new image that combines the foundational structure of the former with the aesthetic qualities of the latter.

In its early phases, style transfer was primarily applied to 2D images~\cite{li2016combining, kolkin2019style, luan2017deep, chen2016fast} and later extended to videos~\cite{huang2017real, xia2021real, li2018learning, 9204808} to achieve temporally consistent stylization across image sequences. Recent works have tackled the 3D style transfer problem, by applying styles to 3D models, such as point clouds~\cite{huang2021learning, mu20223d} and meshes~\cite{hollein2022stylemesh, Liu:Paparazzi:2018}. They stand apart from 2D methods, aiming to ensure a cohesive style across multiple camera angles and enabling free-viewpoint rendering. 
Due to the error-prone geometry stemming from the required 3D reconstruction of them, the stylization of radiance fields~\cite{mildenhall2020nerf} has been actively explored. Methods have incorporated global~\cite{nguyen2022snerf} and local~\cite{zhang2022arf} constraints, utilized stylized reference views~\cite{zhang2023refnpr}, and enhanced diversity through hash encoding and semantic matching~\cite{pang2023locally}. Zero-shot approaches~\cite{liu2023stylerf} have also been developed to circumvent tedious optimization.

These works focus on transferring aesthetic qualities in terms of colors, texture, and brushstrokes from style images to enhance stylization, effectively applying these attributes to 3D scenes. However, the potential benefits of \emph{geometry} remain largely unexplored and neglected. Even though 3D scenes and objects naturally possess both shape and color attributes, most techniques focus solely on color, leaving the geometric parameters unchanged during the style transfer. \citet{nguyen2022snerf} adjust geometry, but the output shapes do not deviate from the original content and fail to reflect geometric cues from the style images.

As noted by art theorists and image creation experts~\cite{kim2020deformable, Metamagical,artandvisual}, geometry has historically played a crucial role in defining and influencing style. The shapes and geometric patterns in an artwork are essential to its stylistic identity.
From this perspective, in the literature of 2D image style transfer, techniques like correspondence search and image warping~\cite{liu2020geometric, Liu_2021_CVPR, kim2020deformable} have been employed to distort image shapes, showcasing how geometry can enhance the expressiveness of stylization.
When applied to 2D images, however, geometric distortion inherently has its limitations. Although shapes are fundamentally 3D forms, images capture only their 2D projections; thus, warping and distorting edges in images offer only an implicit sense of the intended style. It is somewhat limited to assert that they accurately reflect the shape of the objects in the style image.

In this study, we primarily focus on the benefits of incorporating geometric deformation into 3D style transfer. Unlike previous approaches, we define ``geometric style'' as a distinct and clear characteristic that truly captures the geometric essence of the style image. 
Our objective is to transfer these intricate forms into the content of a 3D scene. To the best of our knowledge, our work is the first in style transfer literature to propose \textit{Geometry Transfer}, which extracts geometric style from a style template using a depth map and then directly stylizes the shape of neural radiance fields.
However, merely replacing the RGB style image from the previous methods with a depth map does not produce the desired results. This issue arises from the intrinsic separation between appearance and shape in the radiance fields representation. When the shape is directly optimized, the resulting colors are not well-aligned with the updated form.
To overcome this challenge, we introduce a novel application of deformation fields~\cite{pumarola2021d} that predicts the offset vector for each 3D point. This ensures a harmonious stylization of both appearance and shape during optimization. Consequently, we demonstrate the potential of stylizing the geometry of a 3D scene using a depth map as a style image.
Beyond this demonstration, we introduce innovative techniques to highlight how stylization can benefit from the incorporation of geometry. 

Building on our geometry transfer, we propose a new 3D style transfer method using an RGB-D pair as the style image, aiming for more expressive stylization that better reflects the given style in terms of both shape and appearance. Toward this goal, we propose geometry-aware matching to enhance the diversity of stylization while preserving local geometry through a patch-wise scheme. Additionally, we introduce a novel style augmentation strategy to bring a richer sense of scene depth.
Our contributions are summarized as follows:
\begin{itemize}
    \item For the first time in style transfer literature, we introduce \emph{Geometry Transfer}, a method that extracts style from a depth map and stylizes the geometry of radiance fields.
    \item We propose a novel usage of deformation fields to ensure coherent stylization of both shape and appearance.
    \item We introduce novel RGB-D stylization techniques, enhancing expressiveness and better reflecting the style by leveraging scene geometries.
    \item Our proposed methods can be seamlessly incorporated into existing Panoptic Radiance Fields~\cite{siddiqui2023panoptic}, enabling partial stylization of scenes for more practical applicability.
\end{itemize}

\begin{figure*}[t!]
    \centering
    \includesvg[width=0.99\linewidth]{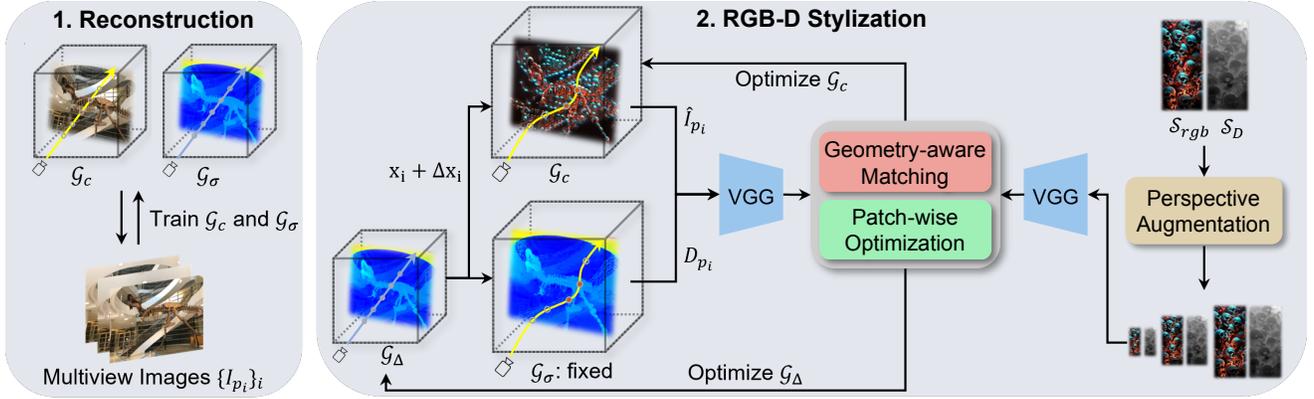}
    \caption{\textbf{Overview of our method.} First we pre-train TensoRF~\cite{chen2022tensorf} on real-world images to obtain the color grid $\mathcal{G}_c$ and density grid $\mathcal{G}_\sigma$, enabling photorealistic reconstruction. Subsequently, we extract VGG features from style images as an RGB-D pair to stylize the shape and appearance of radiance fields. Here, the shape is modified through the additional deformation grid $\mathcal{G}_\Delta$, while $\mathcal{G}_\sigma$ remains fixed.}
    \label{fig:overview}
    \vspace{-0.35cm}
\end{figure*}
\section{Related Works}
\vspace{-1mm}

\label{sec:related_works}
\noindent\textbf{Neural style transfer.} Neural style transfer is the process of creating a new image that fuses the structural elements of a content image with the aesthetic characteristics of a style image. \citet{Gatys2016image} described the style transfer as an iterative optimization that aligns feature correlations from both images, using a deep feature network~\cite{simonyan2015very}. 
Building on this, various techniques~\cite{li2016combining, kolkin2019style, chaudhuri2021semi, luan2017deep, chen2016fast,Kotovenko_2019_ICCV,Kotovenko_2019_CVPR,Kotovenko_2021_CVPR} have advanced stylization through semantic correspondence~\cite{liao2017visual, Zhang_2020_CVPR, huang2019style}, image blending~\cite{tsai2017deep,zhang2020deep,luan2018deep, ke2022harmonizer, cao2023painterly} and novel loss formulas for feature statistics computation~\cite{heitz2021sliced, mechrez2018contextual, risser2017stable}. 
To address the slow convergence of iterative optimization, a feed-forward network has been widely adopted to facilitate arbitrary stylization~\cite{park2019arbitrary, deng2020arbitrary, huang2017arbitrary, gu2018arbitrary, li2017universal, liu2021adaattn, sheng2018avatar} in real-time~\cite{xia2021real, xia2020joint, 9711417, johnson2016perceptual, Sanakoyeu_2018_ECCV}. 
In response to inconsistent stylization across multiple images, techniques to stylize videos~\cite{huang2017real, xia2021real, li2018learning, 9204808, Fruehstueck2023VIVE3D, wang2020Consistent, ruder2018artistic, chen2017coherent} and stereo images~\cite{gong2018neural, chen2018stereoscopic} have been proposed.
Given that shape and geometry are essential for expressive stylization, certain methods have focused on distorting the content's structure, specifically for faces~\cite{yaniv2019face} and text~\cite{Yang2019Controllable}. More general methods~\cite{kim2020deformable,liu2020geometric,Liu_2021_CVPR} utilize correspondence searches and image warping to align content and style from the same class identity. Since shapes and geometry are fundamentally 3D forms, however, modifications to 2D images often fail to capture the accurate style of geometry. Our proposed method aims to directly stylize the shape of the 3D scene using the estimated depth map from the style image.

\noindent\textbf{3D style transfer.}
Recent techniques have applied style transfer to 3D models to ensure coherent stylization across images rendered from multiple viewpoints. Earlier methods stylized explicit representations, such as point clouds~\cite{huang2021learning, mu20223d}, and mesh~\cite{hollein2022stylemesh, Liu:Paparazzi:2018}. More recent techniques~\cite{liu2023stylerf,nguyen2022snerf,zhang2022arf,zhang2023transforming,fan2022unified,pang2023locally,chiang2022stylizing,zhang2023refnpr} have actively explored the stylization of implicit representations, i.e. radiance fields~\cite{mildenhall2020nerf}. In optimization-based approaches~\cite{fan2022unified,pang2023locally}, \citet{nguyen2022snerf} alternated between rendering and 2D stylization, using a global style loss to stylize the 3D scene. Meanwhile, \citet{zhang2022arf, zhang2023refnpr} employed a nearest-neighbor matching loss~\cite{kolkin2022neural} and utilized a reference stylized view~\cite{zhang2023refnpr} to enhance detail preservation. Another direction avoids per-style optimization and instead adopts hypernetworks~\cite{chiang2022stylizing}, feature transformations~\cite{liu2023stylerf}, and Lipschitz mappings~\cite{zhang2023transforming} to facilitate arbitrary stylization of 3D scenes.
While most techniques prioritize appearance without altering geometry during style transfer, we emphasize geometry distortion to improve stylization expressiveness and style accuracy. To our knowledge, this is the first use of a depth map as a style guide to optimize the radiance fields' geometry.
Instead of using reference images for style as above, several approaches stylize radiance fields using text prompts via CLIP embedding~\cite{wang2023nerf,wang2022clip} and leverage diffusion models~\cite{instructnerf2023,kamata2023instruct}. 

\noindent\textbf{Deformation fields.}
Deformation fields have been widely used initially to model the target 3D shape of objects while preserving their geometric details~\cite{jung2022deep,wang20193dn,deng2021deformed}. They define the shape as a surface deformation of the template 3D models. \citet{pumarola2021d} introduced D-NeRF, which employs a time-varying deformation function to capture the transformation between canonical and deformed scenes. This approach allows for the reconstruction of dynamic scenes using a single moving camera. Building on this concept, subsequent studies~\cite{park2021nerfies,peng2021animatable,tretschk2021non, gao2021dynamic} have addressed the view synthesis challenge in dynamic scenes. Our approach distinguishes itself by using deformations to ensure alignment of shape and appearance when stylizing the geometry of the radiance fields.

\section{Methodology}
\label{sec:methods}
\vspace{-1mm}

\noindent\textbf{Preliminaries: Stylizing Radiance Fields.} Stylizing radiance fields is conceptualized as a fine-tuning process that begins with a pre-trained NeRF on a real-world 3D scene. We use TensoRF~\cite{chen2022tensorf} as our scene representation. It introduces two separate grids, $\mathcal{G}_c$ and $\mathcal{G}_\sigma$, each with per-voxel multi-channel features where the former models appearance, and the latter the volume density. To ensure efficient rendering and compact representation, TensoRF factorizes them into multiple low-rank components. For pre-training on the target 3D scene, which includes training images $\{I_i\}^N_{i=1}$ and their corresponding camera poses $\{p_i\}^N_{i=1}$, we follow the training scheme outlined in the original paper and refer the reader there for additional details.

We primarily follow the methods of stylizing radiance fields in ARF~\cite{zhang2022arf}. In each stylization iteration, we randomly select a viewpoint $p_i$ and render the image $\hat{I}_{p_i}$. We then extract 2D feature maps $F^{rgb}_\mathcal{I}$ from $\hat{I}_{p_i}$ and $F^{rgb}_\mathcal{S}$ from the style image, $\mathcal{S}_{rgb}$, using VGG~\cite{simonyan2015very}. After this, we compute the style loss $L_{style}$ between these feature maps, formulated as a nearest-neighbor matching loss~\cite{zhang2022arf, kolkin2022neural}:

\begin{align}
L_{style}=\frac{1}{N}\sum_{i,j}{\min\limits_{i',j'}D(F^{rgb}_\mathcal{I}(i,j),F^{rgb}_\mathcal{S}(i',j'))},
\label{eq:style_loss}
\end{align}
\noindent where $D(,)$ computes the cosine distance between two normalized feature vectors.

\subsection{Geometry Transfer}
Our approach stems from a fundamental question: Can we transfer \textit{geometry} in the same manner as we transfer colors? To explore this, we introduce the use of a depth map as a style guide to transfer its geometry to a 3D scene. An overview of our approach is depicted in Fig.~\ref{fig:overview}.

\subsubsection{Depth Map as a Style Guide}
\label{sec:3.2.1}
\begin{figure}[t]
    \centering
    \includesvg[width=1.0\linewidth]{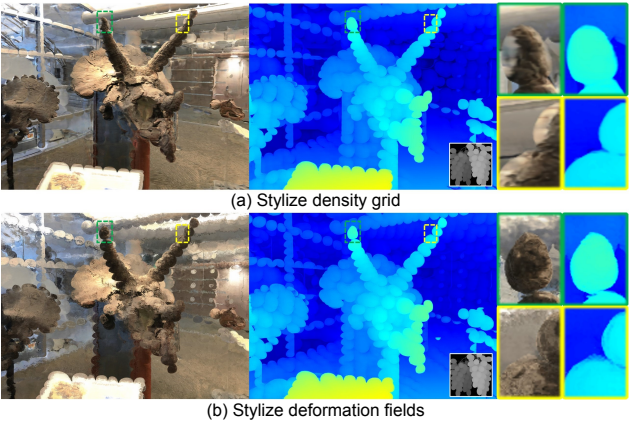}
    \caption{\textbf{Comparisons of the stylized results} obtained by optimizing the density grid (a), and by optimizing the deformation fields (b). When directly optimizing the density, background colors are assigned to the updated parts of the foreground object.}
    \label{fig:deformation}
    \vspace{-0.5cm}
\end{figure}

Instead of using an RGB image as the style guide, we replace it with a depth map, denoted as $\mathcal{S}_{D}$, which captures a distinct style of shape. During the style transfer process, we render the depth map $D_{p_i}$ and optimize the style loss between $D_{p_i}$ and $\mathcal{S}_{D}$. Since the VGG network expects 3-channel images as input, we concatenate the depth maps along the channel dimensions by replicating them three times. Given that $D_{p_i}$ relates solely to volume density, the loss function optimizes the density grid, $\mathcal{G}_\sigma$.  As illustrated in Fig.~\ref{fig:deformation} (a), this approach revealed that we could manipulate shapes in the same manner that we apply style transfer to colors.
However, a challenge arises: while the shape adapts to the style image, the color fields remain static, leading to undesired outcomes. For instance, background colors might be applied to updated portions of foreground objects, even though ideally, the colors of these objects should evolve cohesively with their shape.

\subsubsection{Modeling Deformation Fields}
\label{sec:3.2.2}
\begin{figure}[t]
    \centering
    \includesvg[width=1.0\linewidth]{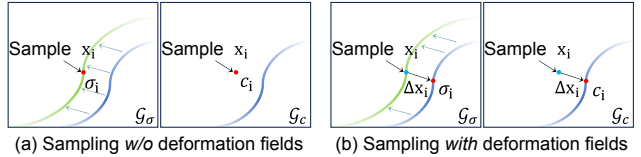}
    \caption{\textbf{Sampling w/ and w/o deformation fields.} Comparisons of the sampling density $\sigma_i$ and color $c_i$ for a 3D point $\mathbf{x}_i$ with and without deformation fields. The curves represent the 2D projected surfaces of objects, where green depicts the stylized surface and blue the original surface. By sampling with deformation fields, we coherently sample both values from the original surface. 
    }
    \label{fig:surface}
    \vspace{-0.25cm}
\end{figure}

After pre-training on real-world scenes, the density grid $\mathcal{G}_\sigma$ forms a surface distribution that mirrors the target 3D scene. Concurrently, color values in the appearance grid $\mathcal{G}_c$ are updated in coherence with the corresponding locations of the surface distribution in $\mathcal{G}_\sigma$. This synchronization leads to the rendering of precise surfaces with accurate appearance.
However, when the geometry is stylized, the surface distribution within $\mathcal{G}_\sigma$ changes, yet $\mathcal{G}_c$ remains consistent. During sampling of 3D points along rays and querying colors and densities from these misaligned grids, the colors of the modified areas are predominantly sourced from the new surface locations in $\mathcal{G}_c$, as shown in Fig.~\ref{fig:surface}~(a), even though the color fields still align with the original distribution.

To address this issue, we introduce a deformation network to enable synchronous modifications of both shape and appearance. This network is designed as a function predicting a three-dimensional displacement vector, $\Delta \mathbf{x}_i\in \mathbb{R}^3$, that maps a 3D point $\mathbf{x}_i$ to its canonical location $\mathbf{x}_i + \Delta \mathbf{x}_i$. In our context, the canonical space refers to the original scene before stylization. We represent the deformation network using another voxel grid, $\mathcal{G}_\Delta$, and update it exclusively for the purpose of stylizing the geometry, keeping $\mathcal{G}_\sigma$ unchanged.
After the stylization, the canonical surface remains intact. When rendering the stylized scene, both the densities and colors are sampled from the original surface locations, as described in Fig.~\ref{fig:surface} (b). This ensures that coherent colors are associated with the modified areas, leading to the differences shown in Fig.~\ref{fig:deformation} (b).

\subsection{RGB-D Stylization}
\vspace{-1mm}

\label{sec:3.3}
To realize a more expressive stylization that modifies both colors and geometry, we employ a pair of style guides: an RGB image and a depth map. Given an RGB style $\mathcal{S}_{rgb}$, we use a zero-shot depth estimation network~\cite{bhat2023zoedepth} to derive its depth map. This then serves as the style depth, $\mathcal{S_D}$.

\subsubsection{Geometry-aware Nearest Matching}
\vspace{-1mm}

To stylize using two style images, specifically $\mathcal{S}_{rgb}$ and $\mathcal{S}_D$, the style loss must be adjusted to account for multiple style sources. Since our goal is to align both colors and geometry, computing the nearest matching loss independently is inappropriate due to potential inconsistencies between patterns of appearance and shape. A more effective method is to initially identify the closest match between content and style features in one domain, then compute the style loss for the other domain using these predetermined pairs. Alternatively, both color and geometry features could be used to search for the nearest neighbors concurrently.
After extracting VGG feature maps from the two modalities, we concatenate them along the channel dimension and then perform a search to find the nearest pair:
\begin{align}
    & {j} = {\arg\min\limits_{i'}D([F^{rgb}_\mathcal{I}(i),F^{D}_\mathcal{I}(i) ],[F^{rgb}_\mathcal{S}(i'),F^{D}_\mathcal{S}(i')])}
\end{align}

We then optimize the cosine distance $D$ separately for features from each modality:
\begin{align}
   & L(i)=D(F^{rgb}_\mathcal{I}(i), F^{rgb}_\mathcal{S}(j)) + D(F^{D}_\mathcal{I}(i),F^{D}_\mathcal{S}(j)),
\end{align}
The style loss is calculated as the mean across all feature vectors: $L_{style}=\frac{1}{N}\sum_{i}{L(i)}$. 
This strategy, which involves incorporating geometry features into the matching process, not only enhances diversity but also better preserves scene structure, as demonstrated in Sec.~\ref{sec:experiments}.

\subsubsection{Patch-wise Optimization}
\vspace{-1mm}

With an RGB style image, it is straightforward to determine if the output aligns with the style in terms of color, texture, and other visual attributes. However, in geometry, depth maps provide limited cues to identify the style. This is because shapes are defined not by isolated pixels, but by their relationship to their surroundings.
The existing nearest matching loss, which conducts matching on a per-pixel basis, is not enough for transferring the style of geometry effectively.
To address this, we introduce a patch-wise matching scheme that broadens the receptive fields, thereby becoming more effective in capturing spatial interactions.

Given the extracted VGG feature maps $F_\mathcal{I}$ and $F_\mathcal{S}$, we first partition each feature map into sets of $k\times k $ patches: $\{\mathcal{P}_\mathcal{I}^i\}_i$ and $\{\mathcal{P}_\mathcal{S}^i\}_i$. The patch-wise style loss $L_\mathcal{SP}$ is then given by:

\begin{align}
L_\mathcal{SP}=\frac{1}{\lvert \mathcal{P}_\mathcal{I}\rvert}\sum_{i}{\min\limits_{j}D^\mathcal{P}(\mathcal{P}_\mathcal{I}^i,\mathcal{P}_\mathcal{S}^j)},
\end{align}

\noindent where $D^\mathcal{P}(\mathcal{P}_1, \mathcal{P}_2)$ computes the sum of the cosine distances between feature vectors at corresponding locations within each patch:
\begin{align}
D^\mathcal{P}(\mathcal{P}_1, \mathcal{P}_2) = \sum_{i}^{k^2}{D(F_1^i, F_2^i)},
\end{align}
Here, $D$ calculates the cosine distance, and 
$F_{1,2}$ represents feature vectors that constitute each patch.
To achieve larger receptive fields without increasing computation,  each patch can be defined with a dilation rate $r$ as a hyperparameter.

\subsubsection{Perspective Style Augmentation}
\vspace{-1mm}

\label{sec:3.3.3} 
We typically select style images with distinct patterns as shown in Fig.~\ref{fig:teaser}, since this aids in the clearer identification of their geometric style. To enhance diversity and the perception of depth, we can vary the sizes of these patterns, applying them differently to surfaces based on their distance.

Before the stylization process, we gather 3D points in world coordinates from all training viewpoints and categorize them into $N$ bins, $\{B_i\}_{i=1}^N$, based on their $z$-coordinates. Each bin $B_i$ is linked to a central value $C_i$, determined by averaging the $z$ values of points within that bin. Given that pattern sizes can vary with the relative resolutions of content and style images~\cite{jing2018stroke}, we modify the style images by downsampling them at multiple scales $\{s_i\}_{i=1}^N$. This process results in a series of style pairs, $\{\mathcal{S}_i\}_{i=1}^N$, where $\mathcal{S} = (\mathcal{S}_{rgb},\mathcal{S}_D)$. We set the scale of the first bin, $s_1$, to $1$. To reflect real-world conditions, the scales of subsequent bins are calculated based on their relative distance from the first bin as: $s_i = C_1/C_i$.

During stylization, each pixel in the rendered image is assigned to a bin $B_{i'}$, based on the shortest distance from the pixel's $z$-coordinate to the center $C_{i'}$ of the bin. This method transforms the rendered image into a format akin to a layered depth image~\cite{LDI}. Each layer is then stylized using its corresponding style pair $\mathcal{S}_{i'}$, which is downsampled to the appropriate scale. Consequently, larger patterns are mapped onto surfaces closer to the viewer, while smaller patterns are applied to more distant surfaces, thereby enhancing the overall sense of depth.

\section{Experiments}
\label{sec:experiments}
\vspace{-1mm}

\begin{figure*}[t!]
    \centering
    \includesvg[width=1\linewidth]{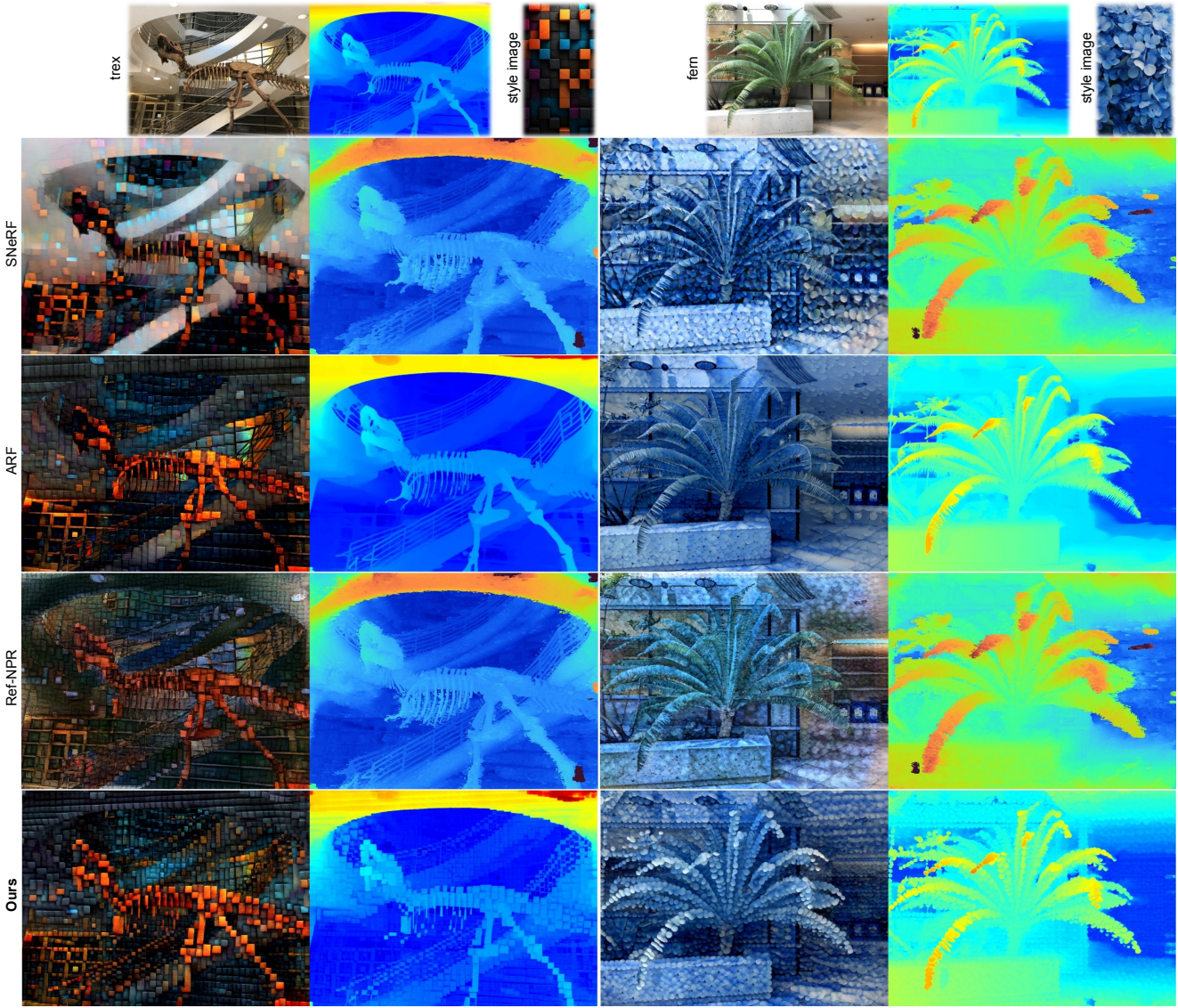}
    \caption{\textbf{Qualitative comparisons} with SNeRF~\cite{nguyen2022snerf}, ARF~\cite{zhang2022arf} and Ref-NPR~\cite{zhang2023refnpr} on the \texttt{trex} and \texttt{fern} scenes~\cite{mildenhall2019local}.}
    \label{fig:comparison}
    \vspace{-0.3cm}
\end{figure*}

\noindent\textbf{Implementation Details.}
We implemented our work based on the code of ARF~\cite{zhang2022arf}, using TensoRF~\cite{chen2022tensorf} as the underlying NeRF representation. For the training of TensoRF during the photorealistic reconstruction stage, we followed the training scheme from its original paper and utilized a distortion regularizer~\cite{barron2022mip} to mitigate artifacts such as floaters and background collapse.
In this stage, the deformation fields are randomly initialized and are optimized to output zeros for all sampled input points.
After pre-training, we maintained the density grid at a constant but updated both the appearance and deformation grids to stylize the reconstructed 3D scene using style loss. We employed the \texttt{conv2} and \texttt{conv3} layers of the VGG-16~\cite{simonyan2015very}, when computing the style loss. We applied the view-consistent color transfer~\cite{zhang2022arf} before and after the stylization. 

\noindent\textbf{Datasets.}
We conducted experiments on the LLFF dataset~\cite{mildenhall2019local}, comprising high-resolution captures of real-world, forward-facing scenes, as used in recent 3D style transfer methods~\cite{zhang2022arf, zhang2023refnpr, nguyen2022snerf}. Furthermore, we utilized the ScanNet dataset~\cite{dai2017scannet} to verify the capability of our approach on scenes captured from diverse camera viewpoints, highlighting our method's potential for partial stylization. The ScanNet dataset includes multiple sequences of real-world indoor scenes characterized by varied trajectories and a collection of common furniture types.

\noindent\textbf{Evaluation metric.}
We use Single Image Fr\'echet Inception Distance (SIFID)~\cite{shaham2019singan} to evaluate the stylizations. SIFID calculates the feature distance between two images, indicating the style similarity between the style image and the stylized results for quantitative evaluation in image style transfer~\cite{wu2022ccpl, pang2021image}. We introduce three methods to assess how both the shape and appearance reflect the specified style guide:
\begin{itemize}
\item \textit{RGB}: To evaluate stylization, we compute the SIFID between the RGB style image, $\mathcal{S}_{rgb}$, and the rendered RGB image, $\hat{I}$.
\item \textit{Gray}: Recognizing that shape and pattern forms, beyond color, influence style, we convert both $\mathcal{S}_{rgb}$ and $\hat{I}$ to grayscale. We then compute the SIFID between these images, allowing us to exclude the influence of color and measure the other style elements.
\item \textit{Depth}: To evaluate the geometry style, we compute the SIFID between the depth style $\mathcal{S}_{D}$ and the rendered depth map, $\hat{D}$.
\end{itemize}

\subsection{Qualitative and Quantitative Comparisons}
In Fig.~\ref{fig:comparison}, we qualitatively compare our results with recent top-performing 3D style transfer methods, including SNeRF~\cite{nguyen2022snerf}, ARF~\cite{zhang2022arf}, and Ref-NPR~\cite{zhang2023refnpr} on the  the \texttt{trex} and \texttt{fern} scenes~\cite{mildenhall2019local}. All these methods stylize radiance fields, guided by a single style image. 
The scale of the style image plays a crucial role in replicating the patterns from the style image; hence, we manually tuned these methods to find the optimal configurations. Since SNeRF did not provide an official implementation, we used an alternative version provided by Zhang \etal~\cite{zhang2023refnpr}, enabling a density update as mentioned in their original paper. For Ref-NPR, we utilized NNST~\cite{kolkin2022neural} to generate a reference stylized view. For ARF, we used the authors' provided TensoRF version since the geometry of the scene is noisy and very inaccurate in the original version with Plenoxels. We applied the distortion regularizer~\cite{barron2022mip} to refine its geometry during pre-training for fair comparisons.

As shown in the figure, our method provides clearer colors and more accurately stylized shapes. Notably, our stylized results replicate the clear and complete forms of style patterns, an ability not achievable by merely stylizing colors, due to the limited space available for mapping complete patterns without altering geometry. To be specific, in order to stylize the \texttt{fern} leaves, it is necessary to change their shape because the leaves are sharp and narrow, which cannot display patterns of the style image without a shape deformation. 
Our method accurately stylizes those regions while the others are limited to stylizing only appearance to just hallucinate the shape. Even though SNeRF updates the density during stylization, the resulting geometry does not reflect any cues from the style image because it lacks proper guidance for geometry style.

In Table~\ref{tab:comparison}, we compare the SIFID~\cite{shaham2019singan} to measure the style similarity between the style images and the rendered images. Our method outperforms others in all metrics, encompassing both appearance and geometry. This demonstrates that incorporating stylization into geometry, as well as colors, enhances the overall style representation and more accurately reflects the intended styles.

\begin{table}
    \centering
        \setlength{\tabcolsep}{1.0mm}
   
        \begin{tabular}{@{}l c c c c c c @{}}
        \toprule
         \multirow{2}{*}{Method}  &  \multicolumn{3}{c}{trex}& \multicolumn{3}{c}{fern} \\
         \cmidrule(lr){2-4} \cmidrule(lr){5-7} 
         &  \textit{RGB} & \textit{Gray} & \textit{Depth} & \textit{RGB} & \textit{Gray} & \textit{Depth} \\
    \midrule
      SNeRF~\cite{nguyen2022snerf}   & 1.62 &	0.81 & 0.59 & 1.32 & 0.64 & 0.40 \\
      ARF~\cite{zhang2022arf}  & 1.54 & 0.64	& 0.51 & 1.11 &	0.48 & 0.36 \\
      Ref-NPR~\cite{zhang2023refnpr}  & 1.59 & 0.72 & 0.61 &1.75 & 0.79 &	0.41\\
      Ours & {\bf 1.43}&{\bf0.58}&{\bf0.44}	&{\bf0.81}	&{\bf0.37}&	{\bf0.28} \\
    \bottomrule
    \end{tabular}
    
    \caption{\textbf{Quantitative comparisons} of SIFID~\cite{shaham2019singan} for RGBs, grayscale images, and depth maps with recent methods. Lower scores indicate better performance. For each scene, images are rendered from $30$ viewpoints, and their average score is computed. }
    \label{tab:comparison}
    \vspace{-0.25cm}
\end{table}

In Table~\ref{tab:user_study}, we present the results of a user study designed to assess visual appeal based on user preferences. We collected rankings from 22 participants for each set of stylization results produced by Ref-NPR~\cite{zhang2023refnpr}, ARF~\cite{zhang2022arf}, and SNeRF~\cite{nguyen2022snerf}, and then computed the average rankings for 12 different stylized scenes. Notably, our proposed method outperforms the others, achieving the highest average ranking. Furthermore, out of 264 total responses ($22\times12$), our method was mostly favored, being selected as the best in 162 instances.

\begin{table}[h]
\centering
\begin{tabular}{@{}c c c c c @{}}
\toprule
metric & Ours & Ref-NPR & ARF  & SNeRF \\
\midrule
Avg. rank $\downarrow$ & {\bf 1.55} & 3.17 & 2.58 & 2.70 \\
\bottomrule
\end{tabular}
\caption{\textbf{User study results} reporting the average ranking.}
    \label{tab:user_study}
    \vspace{-0.25cm}
\end{table}

\subsection{Ablation Experiments}

\begin{figure}[t]
    \centering
    \includesvg[width=1\linewidth]{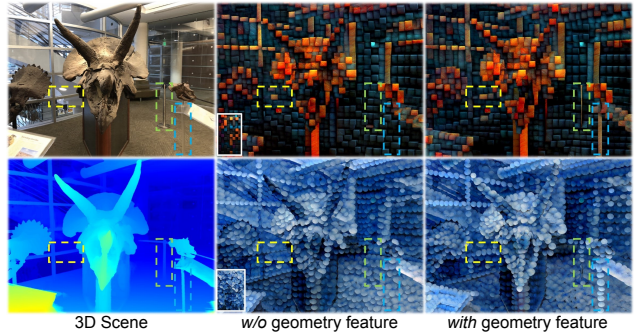}
    \caption{\textbf{Ablation study on the impact of geometric features.} Comparison of the results using nearest matching based on color features versus geometry-aware matching. Geometry features enhance diversity and enable distinct stylizations, differentiating objects with similar colors.}
    \label{fig:gnn}
    \vspace{-0.35cm}
\end{figure}

\noindent\textbf{Geometry-aware nearest matching.} 
In Fig.~\ref{fig:gnn}, we compare the results of the nearest matching exclusively with the color features extracted from $\mathcal{S}_{rgb}$ and our proposed geometry-aware strategy. When using only color features, the resulting style includes similar colors and overlapping patterns in regions with a similar appearance. 
This approach tends to wash out object boundaries, rendering them indistinguishable, and leads to a loss of content structure and diversity, particularly in semi-transparent objects. 
In contrast, when geometry features are also used, the combined consideration of shape and color during the matching process results in distinct colorization and patterns across objects, even those with similar appearances. This differentiation of boundaries enhances content preservation. %

\begin{figure}[t]
    \centering
    \includesvg[width=1\linewidth]{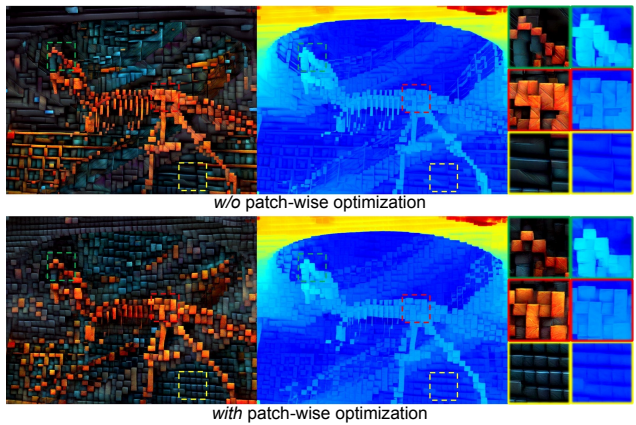}
    \caption{\textbf{Impact of patch-wise optimization}. 
    The patch-wise scheme enhances the clarity and accuracy of patterns and shapes.}
    \label{fig:patch}
    \vspace{-0.4cm}
\end{figure}

\noindent\textbf{Patch-wise optimization.} 
In Fig.~\ref{fig:patch}, we compare the results of the nearest neighbor loss with and without our proposed patch-wise optimization. Without the patch-wise scheme (the top figure), each feature vector in the content and style feature maps is independently matched based on the respective cosine distances, which leads to a failure in maintaining local geometry. Due to its small receptive fields, the scene often contains only incomplete parts of patterns and shapes, resulting in decreased style accuracy. In contrast, when applying the patch-wise optimization (the bottom figure), the positions of local neighbors within the feature maps are preserved during the matching process, enabling the capture of larger receptive fields. This approach results in the intact and complete reproduction of patterns from the style image.

\begin{figure}[t]
    \centering
    \includesvg[width=1\linewidth]{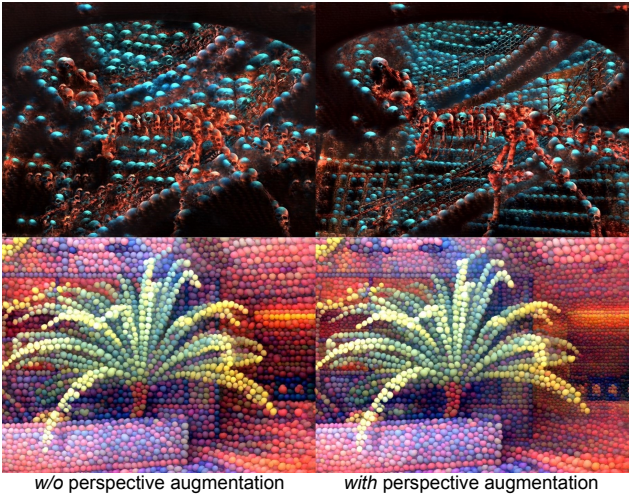}
    \caption{\textbf{Perspective style augmentation impact}. 
    The proposed augmentation enhances depth perception by mapping larger patterns to closer surfaces and smaller patterns to more distant ones.}
    \label{fig:perspective}
    \vspace{-0.4cm}
\end{figure}

\noindent\textbf{Perspective style augmentation.} 
In Fig.~\ref{fig:perspective}, we compare the effects of our proposed perspective style augmentation on stylization. As depicted on the left, stylizing the entire surface with patterns of uniform size, regardless of the distance from the camera, violates perspective rules and diminishes the sense of depth. Conversely, the right column figures demonstrate that decreasing the pattern sizes based on their depth location enhances the perception of depth in the 2D rendered image and aids in preserving the scene's structure.

\subsection{Application: Partial Stylization}

\begin{figure}[t]
    \centering
    \includesvg[width=1\linewidth]{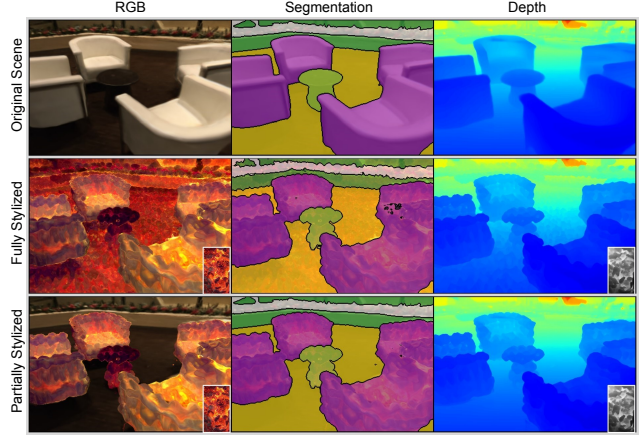}
    \caption{\textbf{Stylization with 3D semantic lifting.} We stylize Panoptic Lifting, pre-trained on the ScanNet dataset~\cite{dai2017scannet}, allowing for the free alteration of target objects for stylization during runtime. As the stylization alters the colors and shapes of objects, the segmentation adapts to their updated forms.}
    \label{fig:panopli}
    \vspace{-0.4cm}
\end{figure}

In Fig.~\ref{fig:panopli}, we demonstrate the applicability of our method in partially stylizing 3D scenes. Instead of partially optimizing the scene based on semantic masks~\cite{lahiri2023s2rf}, we have integrated our method with Panoptic Lifting~\cite{siddiqui2023panoptic}. This approach involves volumetric representations that produce view-consistent panoptic segmentations, along with RGB values and shapes. Our method can be seamlessly incorporated into it, enabling us to dynamically render and select target classes and object instances during \emph{runtime}.

The Panoptic Lifting models a function that maps a 3D point $\mathbf{x}_i$ to color $c_i$, volume density $\sigma_i$, semantic class probability $\kappa_i$, and object id distribution $\pi_i$ over each class as: $\kappa_i(k)\pi_i(j)$. The underlying representation adopts TensoRF and consists of a color grid $\mathcal{G}_c$ and $\mathcal{G}_\sigma$. 
To stylize Panoptic Lifting, we introduce an additional deformation grid $\mathcal{G}_\Delta$ and apply our proposed RGB-D stylization methods to optimize both $\mathcal{G}_c$ and $\mathcal{G}_\Delta$. After the style transfer, we obtain the stylized color grid $\mathcal{G}_{c'}$ and use it to freely render the stylized scene. It is important to note that even if the shape changes after stylization, our use of deformation fields enables sampling from the canonical space for color and density, as well as for the classes and object ids. Thus, if the stylization alters the original shapes, the semantic predictions cohesively adapt to the new shapes.

To render the partially stylized view for specific target classes or objects, we begin by estimating the target class for each 3D point along the rays. During RGB rendering, color is sampled for the 3D points that comprise the target objects from the stylized grid $\mathcal{G}_{c'}$. This sampling is conducted after applying deformation to the points, denoted as $\mathbf{x}_i + \Delta\mathbf{x}_i$. The rest of the scene is rendered using colors from the original grid $\mathcal{G}_c$, with no deformation applied.

\noindent\textbf{Limitations and future work.}
Our selection of TensoRF~\cite{chen2022tensorf} as the underlying representation inherently constrains our capabilities in handling $360^\circ$ unbounded scenes. 
Also, additional challenges arise due to our focus on accurately transferring the shapes and patterns from a single style image to the 3D scene. This task is highly ill-posed as the patterns in 3D scenes do not appear identical when viewed from significantly different perspectives. To effectively stylize $360^\circ$ scenes, it would be beneficial to investigate the use of multi-view style guides or 3D style guides, extending beyond a single-image style reference.

\vspace{0.15cm}
\section{Conclusion}
We proposed Geometry Transfer, a novel method that uses a depth map as a stylistic guide for modifying the geometry of radiance fields. By innovatively employing deformation fields, we achieved coherent alteration of both shape and appearance in 3D scenes. Building upon this foundation, we developed novel RGB-D stylization techniques, leveraging geometric cues to enhance aesthetic expressiveness and more accurately reflect intended styles. Extensive experiments have shown that our methods facilitate a broader spectrum of stylizations compared to previous approaches, significantly expanding the scope of 3D style transfer.

\label{sec:conclusion}

{
    \small
    \bibliographystyle{ieeenat_fullname}
    \bibliography{References.bib}
}

\end{document}